\documentclass[11pt,a4paper]{article}
\usepackage[hyperref]{acl2021}
\usepackage{times} 
\usepackage{latexsym}

\usepackage{microtype}
\usepackage{multirow} 
\usepackage{multicol}
\usepackage{booktabs}
\usepackage{bm}
\usepackage{amsmath}
\usepackage{url}
\aclfinalcopy % Uncomment this line for the final submission
\title{Domain-Adaptive Pretraining Methods for Dialogue Understanding}

\author{Han Wu$^{1}$, Kun Xu$^{2}$, Linfeng Song$^{2}$, Lifeng Jin$^{2}$, Haisong Zhang$^{2}$, Linqi Song$^{1}$\\
         $^{1}$Department of Computer Science, City University of Hong Kong \\ 
	 $^{2}$Tencent AI Lab \\
	 \{\texttt{hanwu32-c}\}\texttt{@my.cityu.edu.hk}\\
	\{\texttt{kxkunxu},\texttt{lfsong},\texttt{lifengjin}, \texttt{hansongzhang}\}\texttt{@tencent.com}\\
	\{\texttt{linqi.song}\}\texttt{@cityu.edu.hk}
}

\date{}

\begin{document}
\maketitle
\begin{abstract}
Language models like BERT and SpanBERT pretrained on open-domain data have obtained impressive gains on various NLP tasks. 
In this paper, we probe the effectiveness of domain-adaptive pretraining objectives on downstream tasks.
In particular, three objectives, including a novel objective focusing on modeling predicate-argument relations, are evaluated on two challenging dialogue understanding tasks.
Experimental results demonstrate that domain-adaptive pretraining with proper objectives can significantly improve the performance of a strong baseline on these tasks, achieving the new state-of-the-art performances.
\end{abstract}

\section{Introduction}

Recent advances in pretraining methods \citep{devlin-etal-2019-bert, joshi2020spanbert, yang2019xlnet} have achieved promising results on various natural language processing (NLP) tasks, including natural language understanding, text generation and question anwsering \citep{liu2019multi, song2019mass, reddy2019coqa}. In order to acquire general linguistic and semantic knowledge, these pretraining methods are usually performed on open-domain corpus, like Wikipedia and BooksCorpus. In light of the success from open-domain pretraining, a further question is naturally raised: whether downstream tasks can also benefit from domain-adaptive pretraining?

To answer this question, later work \citep{baevski-etal-2019-cloze, gururangan-etal-2020-dont} has demonstrated that continued pretraining on the unlabeled data in the target domain can further contribute to the corresponding downstream task.
However, these studies are dependent on additional data that can be unavailable in certain scenarios, and they only evaluated on easy downstream tasks.
For instance, \citet{gururangan-etal-2020-dont} perform continued pretraining with masked language modeling loss on several relevant domains, and they obtain improvements on eight well-studied classification tasks, which are too simple to exhibit the strength of continued domain-adaptive pretraining.
Besides, it is still unclear which pretraining objective is the most effective for each downstream task.

In this work, we give a deeper analysis on how various domain-adaptive pretraining methods can help downstream tasks.
Specifically, we continuously pretrain a BERT model \citep{devlin-etal-2019-bert} with three different kinds of unsupervised pretraining objectives on the domain-specific training set of each target task.
Two of them are Masked Language Model (MLM) \citep{gururangan-etal-2020-dont} and Span Boundary Objective (SBO) \citep{joshi2020spanbert}, both objectives have been explored in previous work.

In addition, a novel pretraining objective, namely Perturbation Masking Objective (PMO), is proposed to better learn the correlation between arguments and predicates.
After domain-adaptive pretraining, the adapted BERT is then tested on dialogue understanding tasks to probe the effectiveness of different pretraining objectives.

We evaluate on two challenging tasks that focus on dialogue understanding, i.e. Conversational Semantic Role labeling (CSRL) and Spoken Language Understanding (SLU).
CSRL \cite{xu2020semantic, xu2021conversational} was recently proposed by extending standard semantic role labeling (SRL) \cite{palmer2010semantic} with cross-utterance relations, which otherwise require coreference and anaphora resolution for being recognized.
We follow previous work to consider this task as sequence labeling.
On the other hand, SLU includes intent detection and slot filling.
To facilitate domain-adaptive pretraining, we only use the training set of each downstream task.
In this way, the usefulness of each pretraining objective can be more accurately examined, as no additional data is used.

Experimental results show that domain-adaptive pretraining significantly helps both tasks.
Besides, our novel objective achieves better performances than the existing ones, shedding more lights for future work on pretraining.

\section{Tasks}
\label{sec:tasks}

\paragraph{Conversational Semantic Role Labeling.}
\citet{xu2021conversational} first proposed the CSRL task, which extends standard SRL by explicitly annotating other cross-turn predicate-argument structures inside a conversation.
Compared with newswire documents, human conversations tend to have more ellipsis and anaphora situations, causing more problems for standard NLU methods.
Their motivation is that most dropped or referred components in the latest dialogue turn can actually be found in the dialogue history.
As the result, CSRL allows arguments to be in different utterances as the predicate, while SRL can only work on each single utterance.
Comparing with standard SRL, CSRL can be more challenging due to the long-range dependencies.
Similar to SRL, we view CSRL as a sequence labeling problem, where the goal is to label each token with a semantic role.

\paragraph{Spoken Language Understanding.}
Proposed by \citet{zhu2020crosswoz}, the SLU task consists of two key components, i.e., intent detection and slot filling.
Given a dialogue utterance, the goal is to predict its intents and to detect pre-defined slots, respectively.
We treat them as sentence-level classification and sequence labeling, respectively.

\section{Domain-Adaptive Pretraining Objectives}
While previous works have shown the benefit of continued pretraining on domain-specific unlabeled data (e.g., \citet{lee2020biobert,gururangan-etal-2020-dont}), these methods only adopt the Masked Language Model (MLM) objective to train an adaptive language model on a single domain.
It is not clear how the benefit of continued pretraining may vary with factors like the objective function.

In this paper, we use the dialogue understanding task as a testbed to investigate the impact of three pre-training objectives to the overall performance.
In particular, we explore the MLM \citep{devlin-etal-2019-bert} and Span Boundary Objective (SBO) \citep{joshi2020spanbert}
, and introduce a new objective, namely Perturbation Masking Objective (PMO), which is more fit for the dialogue NLU task.

\subsection{Masked Language Model Objective}
Masked Language Model (MLM) is the task of predicting missing tokens in a sequence from their placeholders.
Specifically, given a sequence of tokens $X = (x_1, x_2,.., x_n)$, a subset of tokens $Y \subseteq X$ is sampled and substituted with a different set of tokens.
In BERT's implementation, $Y$ accounts for 15\%
of the tokens in $X$; of those, 80\% are replaced with [\texttt{MASK}], 10\% are replaced with a random token (according to the unigram distribution), and 10\% are kept unchanged.
Formally, the contextual vector of input tokens $X$ is denoted as $\bm{H} = (\bm{h}_1, \bm{h}_2,..., \bm{h}_n)$.
The task is to predict the original tokens in $Y$ from the modified input and the objective function is:
\begin{equation} \label{equ:mlm_loss}\nonumber
\mathcal{L}_{MLM} = - \frac{1}{|Y|} \sum_{t=1}^{|Y|} \log p(x_t|\bm{h}_t;\bm{\theta})
\end{equation}
where $|Y|$ is the number of masked tokens, and $\bm{\theta}$ represents the model parameters.

\subsection{Span Boundary Objective}
In many NLP tasks such as the dialogue understanding, it usually involves reasoning about relationships between two or more spans of text. 
Previous works \cite{joshi2020spanbert} have shown that
SpanBERT is superior to BERT in learning span representations,
which significantly improves the performance on those tasks.
Conceptually, the differences between these two models are two folds.

Firstly, different with BERT that independently selects the masked token in $Y$, SpanBERT define $Y$ by randomly selecting \textit{contiguous spans}. In particular, SpanBERT first selects a subset $Y \subseteq X$ by iteratively sampling spans until masking 15\% tokens\footnote{The length of each span is sampled from the geometric distribution $l \sim Geo(p)$, with $p=0.2$.}. Then, it randomly (uniformly) selects the starting point for the span to be masked.

Secondly, SpanBERT additionally introduces a span boundary objective that involves predicting each token of a masked span using only the representations of the observed tokens at the boundaries. For a masked span of tokens $(x_s,..., x_e) \in Y$, where $(s, e)$ are the start and end positions of the span, it represents each token in the span using the boundary vectors and the position embedding:
\begin{equation}\nonumber
    \bm{y_i} = f(\bm{h}_{s-1}, \bm{h}_{e+1}, \bm{p}_{i-s+1})
\end{equation}
where $p_{i}$ marks relative positions of span token $x_i$ with respect to the left boundary token $x_{s-1}$, and $f(\cdot)$ is a 2-layer MLP with GeLU activations and layer normalization.
SpanBERT sums the loss from both the regular MLM and the span boundary objectives for each token in the masked span:
\begin{equation} \label{equ:span_loss} \nonumber
\mathcal{L}_{SBO} = - \frac{1}{|Y|} \sum_{t=1}^{|Y|} \log p(x_t|\bm{y}_t;\bm{\theta})
\end{equation}

\subsection{Perturbation Masking Objective}
In dialogue understanding tasks like CSRL, the major goal is to capture the semantic information such as the correlation between arguments and predicate.
However, for the sake of generalization, existing pretraining models do not consider the semantic information of a word and also not assess the impact of predicate has on the prediction of arguments in their objectives.
To address this, we propose to use the perturbation masking technique \cite{wu-etal-2020-perturbed} to explicitly measure the correlation between arguments and predicate and further introduce that into our objective.

The perturbation masking is originally proposed to assess the impact one word has on the prediction of another in MLM.
In particular, given a list of tokens $X$, we first use a pretrained language model \textbf{M} to map each $x_{i}$ into a contextualized representation $H(X)_i$.
Then, we use a two-stage approach to capture the impact word $x_j$ has on the prediction of another word $x_i$.
First, we replace $x_i$ with the [\texttt{MASK}] token 
and feed the new sequence $X\backslash\{x_i\}$ into \textbf{M}.
We use $H(X$$\setminus$$\{x{_i}\})_{i}$ to denote the representation of $x_i$.
To calculate the impact $x_j\in x$$\setminus$$\{x_i\}$ has on $H(X)_i$, we further mask out $x_j$ to obtain the second corrupted sequence $X$$\setminus$$\{x_i, x_j\}$.
Similarly, $H(X$$\setminus$$\{x{_i}, x_{j}\})_{i}$ denotes the new representation of token $x_i$.
We define the the impact function as: 
$f(x_i, x_j) = d(H(X$$\setminus$$\{x{_i}\})_{i}, H(X$$\setminus$$\{x{_i}, x_{j}\})_{i})$,
where $d$ is the distance metric that captures the difference between two vectors.
In experiments, we use the Euclidean distance as the distance metric.

Since our goal is to better learn the correlation between arguments and predicate,
we introduce a perturbation masking objective that maximizes the impact of predicate on the prediction of argument span:
\begin{equation} \label{equ:pmo_loss} \nonumber
    \mathcal{L}_{PMO} = - \frac{1}{|Y|} \sum_{t=1}^{|Y|} - f(x_{t}, \{x_{p_{0}}, ..., x_{p_{m-1}}\})_{i}
\end{equation}
where $p_{0}$,... $p_{m-1}$ are $m$ predicates that occur in the sentence.
In practice, we first follow the SpanBERT to sample a subset of contiguous span texts and perform masking (i.e., span masking) on them.
Then, we select verbs from $X$ as predicates and perform perturbation masking on those predicates.

\begin{table*}[ht!]
\small
\fontsize{10}{11} \selectfont
\setlength\tabcolsep{4pt}
\centering
\bgroup
\def\arraystretch{1.2}
\begin{tabular}{lcccccccccccc}
\toprule[0.8pt]
\multirow{2}{*}{Pretraining Strategy} & \multicolumn{3}{c}{DuConv} & & \multicolumn{3}{c}{NewsDialog} & & \multicolumn{3}{c}{CrossWOZ} \\
\cline{2-4} \cline{6-8} \cline{10-12}
& F1$_{all}$ & F1$_{cross}$ & F1$_{intra}$ & & F1$_{all}$ & F1$_{cross}$ & F1$_{intra}$ & & F1$_{intent}$ & F1$_{slot}$ & F1$_{all}$ \\
\hline
No Pretraining  & 88.16 & 83.74 & 88.71 & & 76.81 & 53.61 & 79.97 & & 95.67 & 95.13 & 95.34 \\
\hline
MLM & 88.56 & 84.37 & 88.97 & & 76.93 & 53.43 & 80.15 & & 95.85 & 95.47 & 95.62 \\
MLM + SBO & 88.73 & 84.49 & 89.23 & & 78.10 & 56.21 & 80.85 & & {96.17} & {95.54} & {95.78} \\
MLM + PMO & 89.10 & 85.26 & 89.52 & & 79.68 & 56.19 & 81.79 & & 96.40 & 95.79 & 96.17 \\
MLM + SBO + PMO &  89.21 & 85.98 & 89.79 & & 80.01 & 56.20 & 82.78 & & 96.48 & 96.03 & 96.21 \\
\hline
\quad w/ NP Sampling ($\alpha=50$)&  89.34 & 86.12 & 89.99 & & 81.32 & 56.67 & 83.14 & & 96.81 & 96.52 & 96.70 \\
\quad w/ NP Sampling ($\alpha=80$)&  \textbf{89.97} & \textbf{86.68} & \textbf{90.31} & & \textbf{81.90} & \textbf{56.56} & \textbf{84.56} & & \textbf{96.97} & \textbf{96.87} & \textbf{96.93} \\
\toprule[0.8pt]
\end{tabular}
\egroup
\caption{Evaluation on the DuConv, NewsDialog and CrossWOZ. $\alpha$ is the ratio of sampling from noun phrases.}
\label{tab:results}
\vspace{-5mm}
\end{table*}

\section{Experiments}
We evaluate pretraining objectives on three datasets, DuConv, NewsDialog\footnote{We obtain the CSRL annotations on DuConv and NewsDialog directly from the author of \citet{xu2021conversational}.} and CrossWOZ.
The former two datasets are annotated by \citet{xu2021conversational} for the CSRL task and the last one is provided by \citet{zhu2020crosswoz} for the SLU task.

Duconv is a Chinese knowledge-driven dialogue dataset, focusing on the domain of movies and stars.
NewsDialog is a dataset collected  in a way that follows the setting for constructing general open-domain dialogues: two participants engage in chitchat, and during the conversation, the topic is allowed to change naturally.
\citet{xu2021conversational} annotates 3K dialogue sessions of DuConv to train their CSRL parser, and directly test on 200 annotated dialogue sessions of NewsDialog.
CrossWOZ is a Chinese Wizard-of-Oz task-oriented dataset, including 6K dialogue sessions and 102K utterances on five domains.

Since the state-of-the-art models on these tasks are all developed based on BERT,
we use the same model architectures but just replace the BERT base with our domain-adaptive pretrained BERT.
Notice that, we also experiment with other pretrained language models such as RoBERTa and XLNet. We observed similar results but here we only report the results based on BERT due to the space limitation.

In particular, we perform the domain-adaptive pretraining on CSRL task using all dialogue sessions of training set in DuConv \citep{wu-etal-2019-proactive} and NewsDialog \citep{wang2021naturalconv}, which includes 26K and 20K sessions, respectively; on the SLU task, we use the whole CrossWOZ training dataset.

The hyper-parameters used in our model are listed as follows.
The network parameters of our model are initialized using the pretrained language model.
The batch size is set to 128. We use Adam \cite{kingma2014method} with learning rate 5e-5 to update parameters.

\noindent\paragraph{Results and Discussion.}
On the CSRL task, we follow \citet{xu2021conversational} to use the micro-averaged F1 over the (\textit{predicate}, \textit{argument}, \textit{label}) tuples. Specifically, we calculate F1 over all arguments (referred as F1$_{all}$) and those in the same and different dialogue turns as predicates (referred as F1$_{intra}$ and F1$_{cross}$).
On the SLU task, we report results on F1$_{intent}$, F1$_{slot}$ and F1$_{all}$.
Table~\ref{tab:results} summarizes the results. 
The first row shows the performance of existing state-of-the-art models without domain-adaptive pretraining on each dataset.
We can see that on two tasks, existing models could benefit from the domain-adaptive pretraining,
achieving new state-of-the-art performance on these datasets.

Let us first look at the CSRL task. Pretraining with MLM objective could slightly improve the performance by 0.4 and 0.12 in terms of F1$_{all}$ on DuConv and NewsDialog, respectively.
By additionally considering the span boundary objective, the overall performance especially F1$_{cross}$ could be further improved by at least 0.75 and 2.6, respectively.
These results are expected since arguments in the CSRL task are usually spans and SBO is better than MLM in learning the span representation.
We can also see that our proposed perturbation masking objective boosts the performance by a larger margin than SBO, indicating that learning correlations between arguments and predicates is more crucial to the NLU task.
By summing three objectives, the CSRL model could achieve the best performance, significantly improving the baseline that without domain-adaptive pretraining by 1.05 and 3.2 F1$_{all}$ score, respectively.

From Table~\ref{tab:results}, we can see that similar findings are also observed on the SLU task.
First of all, domain-adaptive pretraining on CrossWOZ could also improve the performance.
Secondly, adding either SBO or PMO, the F1 scores on intent and slot could be further improved.
Thirdly, the best performance is achieved when all three objectives are considered.
However, we do not observe similar substantial gains on the SLU task as on the CSRL task.
We think this is because the state-of-the-art performance on CrossWOZ is relatively high, but it is still impressive to achieve absolute 0.81, 0.90 and 0.87 points improvement in terms of F1$_{intent}$, F1$_{slot}$ and F1$_{all}$.

We also investigate the impact of span masking scheme to the overall performance.
Recall that, in the span masking, we randomly sample the span length and a start position of the span.
\citet{joshi2020spanbert} showed that no significant performance gains are observed by using more linguistically-informed span masking strategies such as masking \textit{Named Entities} or \textit{Noun Phrases}.
Specifically, they use the spaCy's\footnote{https://spacy.io/} named entity recognizer and constituency parser to extract named entities and noun phrases, respectively.
In this paper, we revisit these span masking scheme. Since there is no available constituency parser designed for the dialogue, we use an unsupervised grammar induction method \cite{jin2020grounded} to extract grammars from the training data.
\textbf{Noun phrases} from Viterbi parse trees from different grammars are tallied without labels,
resulting in a posterior distributions of the spans, which are used in our span sampling.
As shown in Table~\ref{tab:results}, we find the best choice is to combine random sampling and noun phrases sampling, i.e., sampling from the noun phrases at $\alpha$\% of the time and from a geometric distribution for the other (1 - $\alpha$\%).
The performance on all three datasets coherently increases when more noun phrases are used in the span sampling.

\section{Conclusion}
In this paper, we probe the effectiveness of domain-adaptive pretraining on dialogue understanding tasks. Specifically, we study three domain-adaptive pretraining objectives, including a novel objective: \textit{perturbation masking objective} on three NLU datasets.
Experimental results show that domain-adaptive pretraining with proper objectives is a simple yet effective way to boost the dialogue understanding performance.

\section*{Acknowledgement}
We would like to thank the anonymous reviewers for their valuable and constructive comments.
This work was supported in part by the City University of Hong Kong Teaching Development Grants 6000755.

\bibliographystyle{acl_natbib}
\bibliography{acl2021}

%\appendix
\end{document}